\documentclass[10pt,twocolumn,letterpaper]{article}
\usepackage[cvpr]{cvpr}      

\usepackage{graphicx}
\usepackage{amsmath}
\usepackage{amssymb}
\usepackage{booktabs}

\usepackage[pagebackref,breaklinks,colorlinks]{hyperref}

\usepackage[capitalize]{cleveref}
\crefname{section}{Sec.}{Secs.}
\Crefname{section}{Section}{Sections}
\Crefname{table}{Table}{Tables}
\crefname{table}{Tab.}{Tabs.}

\def\cvprPaperID{*****} 
\def\confName{CVPR}
\def\confYear{2022}

\begin{document}

\title{Visual Watermark Removal Based on Deep Learning}
\author{Rongfeng WEI\\
City University of Hong Kong\\
{\tt\small rfwei2@my.cityu.edu.hk}
}
 
\maketitle
\begin{abstract}
    In recent years as the internet age continues to grow, sharing images on social media has become a common occurrence. In certain cases, watermarks are used as protection for the ownership of the image, however, in more cases, one may wish to remove these watermark images to get the original image without obscuring. In this work, we proposed a deep learning method based technique for visual watermark removal. Inspired by the strong image translation performance of the U-structure, an end-to-end deep neural network model named AdvancedUnet is proposed to extract and remove the visual watermark simultaneously. On the other hand, we embed some effective RSU module instead of the common residual block used in UNet, which increases the depth of the whole architecture without significantly increasing the computational cost. The deep-supervised hybrid loss guides the network to learn the transformation between the input image and the ground truth in a multi-scale and three-level hierarchy. Comparison experiments demonstrate the effectiveness of our method. 
\end{abstract}

\section{Introduction}
\label{sec:intro}

Watermark removal is an open and challenging problem with the aim of reconstructing the background image on top of the watermarked image. Watermarks can be overlaid anywhere on a background image of varying size, shape, color and transparency. Furthermore, watermarks often contain complex patterns such as distorted symbols, thin lines, shadow effects, and removing these watermarked images and restoring the original image is a challenging task. The structure, position and size of these watermarks vary from image to image and without manual guidance or assumptions about the underlying image, it will be difficult to detect the watermarked image and reconstruct the original image.

With the continuous development of GPU in recent years, the computing power of computers has also increased significantly, thus promoting the rapid progress of deep learning methods and generating a series of techniques based on convolutional neural networks. Through continuous convolution and pooling operations, they can extract multi-level contextual high-level features with rich semantics and learn the potential universal laws of images, thus far surpassing traditional methods.

In summary, image watermark removal as a branch of image processing tasks has a very meaningful application value and social value, in today's era of the prosperous information society, the development of watermark removal technology also continues to advance the adversarial progress of watermark technology, which  is significant to the field of network information security. Compared with traditional methods, deep learning methods have stronger robustness, and the methods are more concise and efficient. In this paper, we use BVMR\cite{BVPR}, introduced in CVPR2019, as a benchmark to improve the "blind" watermark removal method based on deep learning and achieve experimental results that significantly exceed the benchmark results. 

Our main contributions could be summarized as follows,  
\begin{itemize}
 
\item We propose a one-stage neural network named AdvancedUnet to extract and remove the visual watermark simultaneously.   

\item In each stage of the U-structure, we embed an effective RSU module instead of the original module in UNet, which enables the network to learn deeper features and be more robust with multi-scale.  

\item A novel deep-supervised-hybrid loss that fuses BCE, SSIM, and IoU in each stage is proposed to predict the mask of watermark more accurately. 

\item Our method surpasses the baseline by a large margin.

\end{itemize}

\section{Related Work}
\label{sec:formatting}

Image watermark removal can be considered as a two-stage multitasking technique for target detection and image reconstruction or as a one-stage image translation task.

\subsection{Saliency detection}

The human vision system has an effective attention mechanism for choosing the most important information from visual scenes. For a curtain task like watermark detection, what attracts human attention most is the watermark, and so does the computer. Furthermore, the detection of watermark should reach the pixel-to-pixel level, which will be the input of background reconstruction. Consequently, we regard the detection of watermark as a task of saliency detection. Qin et al.\cite{BASNet} propose a novel boundary-aware salient object detection network: BASNet, which consists of a deeply supervised encoder-decoder and a residual refinement module. And a novel hybrid loss that fuses BCE, SSIM, and IoU is proposed to supervise the training process of accurate salient object prediction on three levels: pixel-level, patch-level, and map-level. Also, Qin et al.\cite{U2-Net} propose a novel two-level nested U-structure, which allows the network to go deeper, attain high resolution, without significantly increasing the memory and computation cost.

\subsection{Image reconstruction }

The main purpose of image inpainting is to produce visually plausible structure and texture for the missing regions of damaged images. Famous works treated watermark removal as an image-to-image translation task and directly map watermarked images to watermark-free ones. 

Cun and Pun\cite{S2AM} start from an empirical observation: the inharmonious appearance can only be found in local inharmonious objects, and they share the same semantic information and the appearance in the background region. Thus, this paper introduces a novel attention module named Spatial-Separated Attention Module for learning the features in the foreground and background area by hard-coded masks individually. \cite{DLinpainting}is a survey that presents a comprehensive overview of recent advances in deep learning-based image inpainting, which gives a detailed analysis on the performance of different inpainting algorithms. Zamir et al.\cite{Multi-Stage} propose a multi-stage architecture, that progressively learns restoration functions for the degraded inputs, thereby breaking down the overall recovery process into more manageable steps. 

\subsection{Watermark removal }

Hertz et al.\cite{BVPR} propose a deep learning-based technique for blind removal of watermark. In the blind setting, the location and exact geometry of the watermark are unknown. This approach simultaneously estimates which pixels contain the visual motif, and synthesizes the underlying latent image. To the best of our knowledge, there are only a few deep learning methods specifically designed for watermark removal, and this work is the only existing algorithm using image-to-image translation method without other reductant modules, which is quite simple and efficient. But in the follow-up research, researchers usually regard the refinement module is indispensable in watermark removal task. Liu et al.\cite{WDNet} design their network as a combination of DecompNet and RefineNet. \cite{SplitthenRefine} and \cite{SCLBR} also propose a two-stage framework to simulate the process of detection, removal and refinement.  

However, the two-stage models must sacrifice speed for accuracy, which are always made up of many complex modules. In this paper, continuing the concise idea in BVMR, we propose a one-stage neural network named AdvancedUnet to extract and remove the visual watermark simultaneously. Experiments demonstrate that our method surpasses the baseline by a large margin. 


\section{Data}
\label{sec:formatting}

In the image watermark removal technique, we need to use the binarization mask of the watermark as the ground truth, which is similar to a watermark segmentation problem requiring very fine pixel-level manual extraction and will be very labor-intensive, so in this paper, we use automatic watermark image synthesis to generate the dataset for character watermarks. 

In the experiment, we synthesize the watermark images by embedding visual watermarks into background images, where the visual watermarks are simplified to be random character strings with different opacities and the background images are random from Microsoft COCO val2014 dataset \cite{COCO}. Matting a watermark $(V_{m})$ onto an image $(I_{m})$ can be obtained by: 
\begin{equation}
 C_{r} = \alpha \circ V_{m} + (1 - \alpha) \circ I_{m}  
\end{equation}
where $C_{r}$ is the synthesized corrupted image, and $\alpha$ is the spatially varying transparency. 

\section{Methods}
\label{sec:formatting}

\subsection{AdvancedUnet}
\begin{figure}[t]
  \centering
   \includegraphics[width=0.8\linewidth]{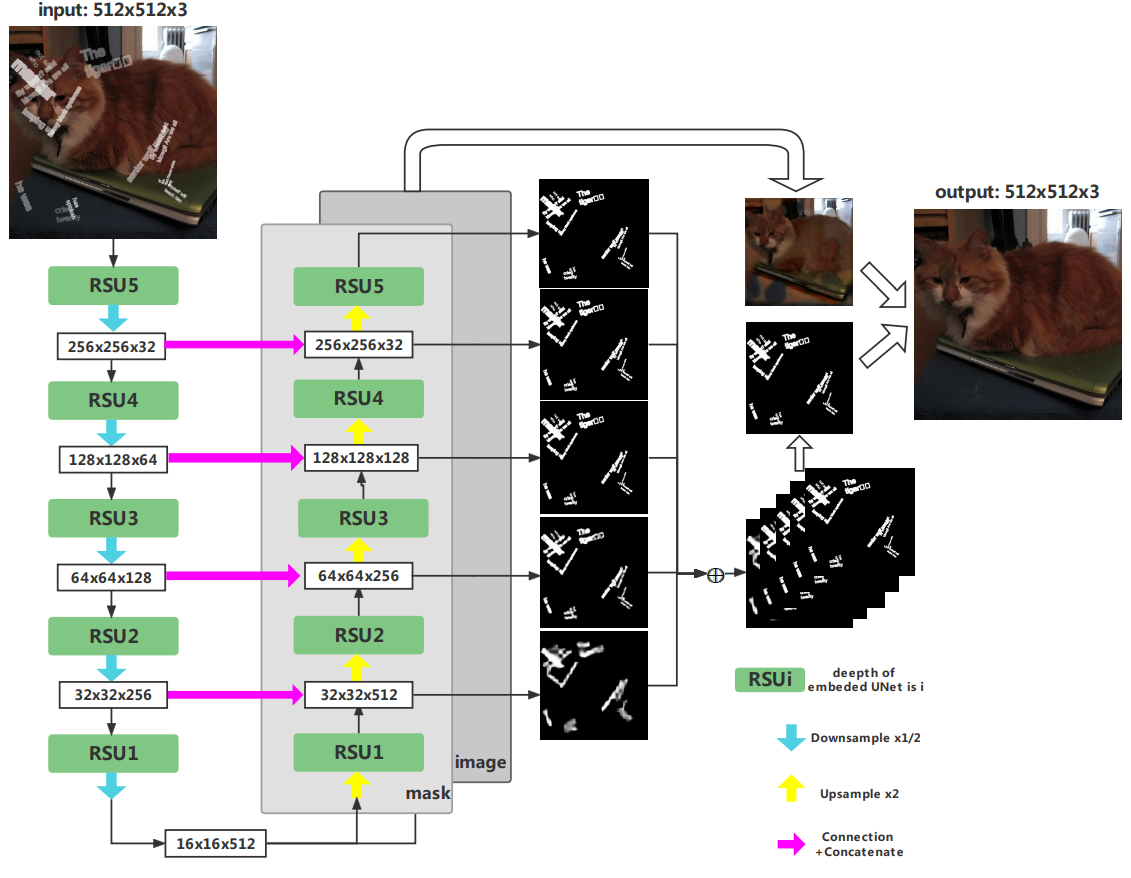}
   \caption{AdvancedUnet}
   \label{fig:Adv}
\end{figure}

Fig.\ref{fig:Adv} shows in detail the network structure of the image watermark removal algorithm model AdvancedUnet we proposed. Specifically, it  bases on a Y-structure network, which has one encode and two decoders, the two decoders will output the binary mask of guessed watermark and the reconstructed non-watermark image respectively. With an input of 512*512 shape synthesized watermark image, the feature map goes through operations such as upsampling, downsampling, jump connection and convolution in the RSU module, and the feature map does not change its size for each RSU module it passes through. In the encoder stage, the down-sampling operation is performed by the RSU module after the output, which makes the feature map halve its length and width and multiply the number of channels. In the decoder stage, after output by the RSU module, the feature map is channel-connected with the coded output of the corresponding encoder stage, and then the upsampling operation is performed. As the binary watermark masks shown in the figure, we will count predicted binary mask from each stage with ground truth by the deep-supervised hybrid loss.  

The final image is reconstructed by replacing all the pixels in the estimated visual watermark region in the original corrupted image $C_{r}$ with the corresponding pixels from the reconstructed image $\widehat{I_{m}}$. The final resulting image is given constructed image $I_{m}$ by 
\begin{equation}
Im_{final} = (1 - \widehat{Ma}) \circ C_{r} + \widehat{Ma} \circ \widehat{I_{m}}
\end{equation}, where $\circ$ denotes element-wise multiplication.


\subsection{RSU Module}
\begin{figure}[t]
  \centering
   \includegraphics[width=0.6\linewidth]{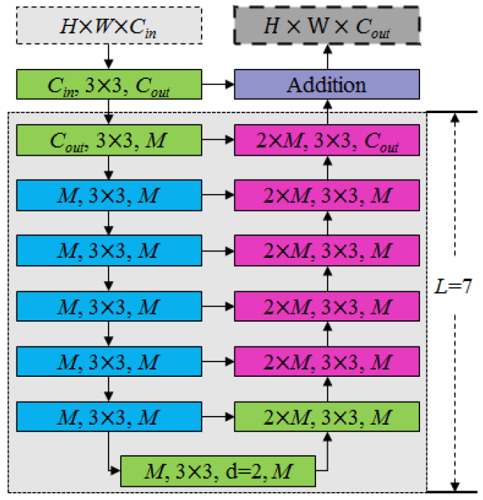}
   \caption{Residual U-block}
   \label{fig:RSU}
\end{figure}

ReSidual U-block, RSU\cite{U2-Net}, to capture intra-stage multi-scale features. The structure of $RSU-L(C_{in}M, C_{out} )$ is shown in Fig.\ref{fig:RSU}, where L is the number of layers in the encoder, $C_{in} , C_{out}$ denote input and output channels, and M denotes the number of channels in the internal layers of RSU. In the architecture of our model, the depth of RSU varies from each stage. This design change empowers the network to extract features from multiple scales directly from each residual block, the RSU is deeper when the resolution of feature map is larger. More notably, the computation overhead due to the U-structure is small, since most operations are applied on the downsampled feature maps. Consequently, we utilize the RSU module to enhance the feature extract capability of our model without increasing too much computational expense. 
\subsection{Deep supervised hybrid loss}
During training, we used a deep supervised approach similar to HED \cite{HED}, where a prediction map is output for each layer of both decoders, the prediction maps are resized to the input map size using bilinear interpolation, and the loss value relative to the real map is computed for prediction map from each stage as well as the fused map of each prediction map. 

Similar to the saliency target detection and shadow detection problems, the detection of watermark masks is also a binarized partitioning problem, so we choose to use binary cross-entropy loss for not only each layer but also a fused feature map of the watermark mask decoder branch to compute the loss of the prediction map. In order to enhance the robustness of model in image structure, we introduce the SSIM loss into our hybrid loss. 

The hybrid loss combines the binary cross entropy (BCE), structural similarity (SSIM) and IoU losses, to supervise the training process in a three-level hierarchy: pixel-level, patch-level and map- level. 

loss for binary mask:  
\begin{equation}
\begin{split}
Loss_{mask}(M) = \sum_{l \in layers}[L^{ssim}(M_{l}) + l^{bce}(M_{l})]\\+L^{ssim}(M_{fuse})+l^{bce}(M_{fuse})
\end{split}
\end{equation}

loss for reconstructed image: 
\begin{equation}
Loss_{image}(M) = \sum_{l \in layers}L^{ssim}(M) + L^{1}(M)
\end{equation}

total loss: 
\begin{equation}
Loss(M) =Loss_{mask}(M) + Loss_{image}(M) 
\end{equation}

\section{Experiments}
\subsection{Implementation details}
We use PyTorch \cite{paszke2019pytorch} with CUDA v11.2 to implement our algorithm. The training batch size equal to 8 under the Tesla P100 GPU on the colab platform. For a fairer comparison, we utilize identical optimizer as the baseline\cite{BVPR}, and our project is based on \cite{BVPR}.

In the experiment, by observing the output loss line graph of decoder at each level during training (as in Fig.3), we find that the fluctuation of $loss_{mask1}$ is the largest, consistent with our algorithm process, the loss fluctuation decreases gradually as the level rises. At the lower level, the feature map size is smaller, and its loss calculation relative to the real image needs to be calculated by upsampling to the same size as the real image. Such up-sampling operation often brings huge accuracy loss, and the compensation information of the encoder jump connection input obtained at lower levels is also less, so it has great unreliability, so we will give a smaller weight to $loss_{mask1}$ in the calculation of the actual experimental loss function.

\subsection{Comparison with baseline}
\subsubsection{Qualitative analysis}
The experimental results can be visualized as Figure.3 and analyzed from two perspectives: the watermark mask and the reconstructed image.

(1) The benchmark model \cite{BVPR} can roughly predict the location of the watermark in the watermark mask prediction. Still, it can not do pixel-level prediction for watermarks with high transparency, and there are very many artifacts, as shown in Figure 4. Our model, on the other hand, can predict each part of the watermark more accurately, and the edge processing is more ideal, but for the above figure, there is a piece of watermark position in the location of the seriously reflected light, making the model prediction disturbed.

(2) The benchmark model is able to remove part of the watermark, but there is still a large amount of watermark residual marks, while the effect of our model is so stunning that it is almost impossible to distinguish our algorithm reconstructed image from the original watermark-free image without a watermarked image comparison.
\begin{table}
  \centering
  \begin{tabular}{c c c c c c}
    \toprule
    Method/Metric & MAE & mIoU & SSIM & PSNR \\
    \midrule
    Baseline & 0.023 & 0.748 & 0.966 & 28.041\\
    Ours & 0.010 & 0.868 & 0.976 & 39.762\\
    \bottomrule
  \end{tabular}
  \caption{Quantitative analysis}
  \label{tab:example}
\end{table}
\begin{figure}[t]
  \centering
   \includegraphics[width=0.8\linewidth]{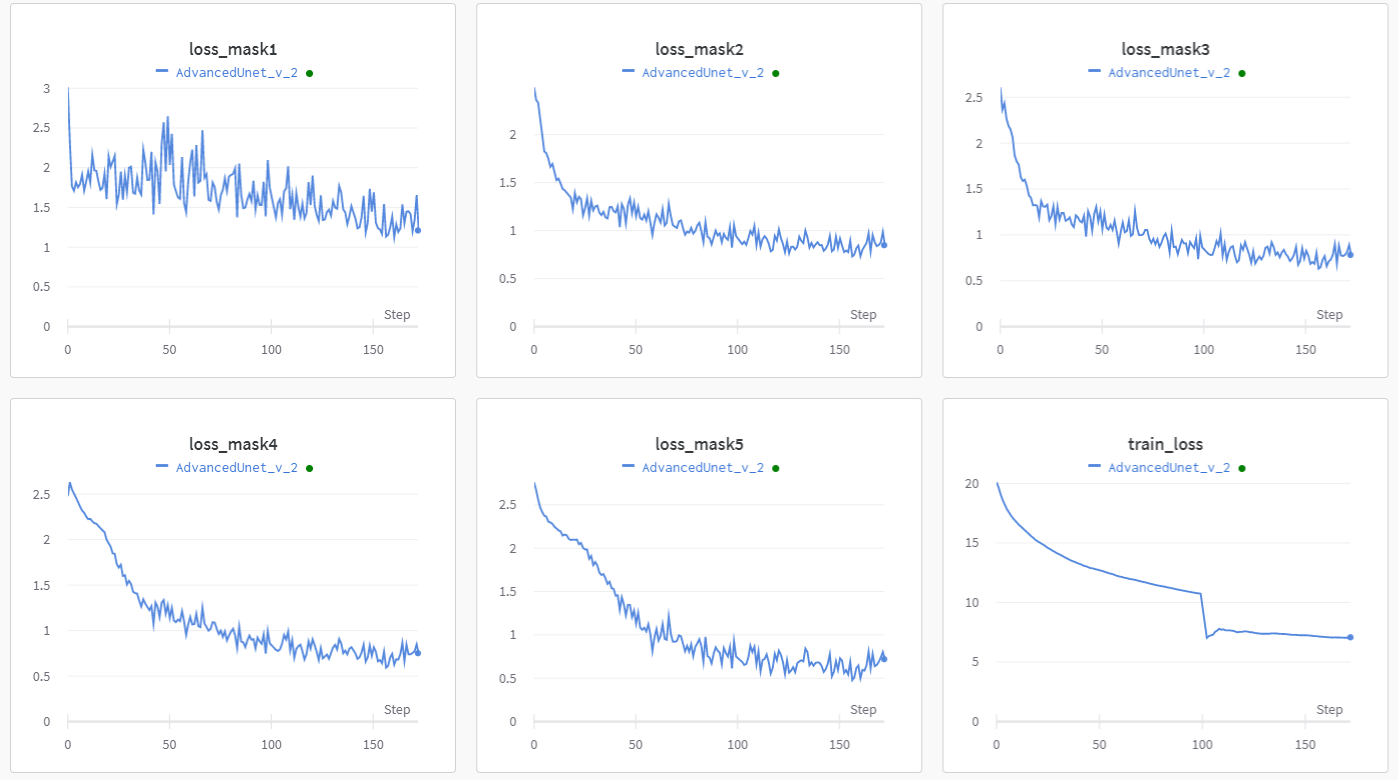}
   \caption{Loss diagram for each level of output mask}
   \label{fig:onecol}
\end{figure}

\begin{figure}[t]
  \centering
   \includegraphics[width=0.7\linewidth]{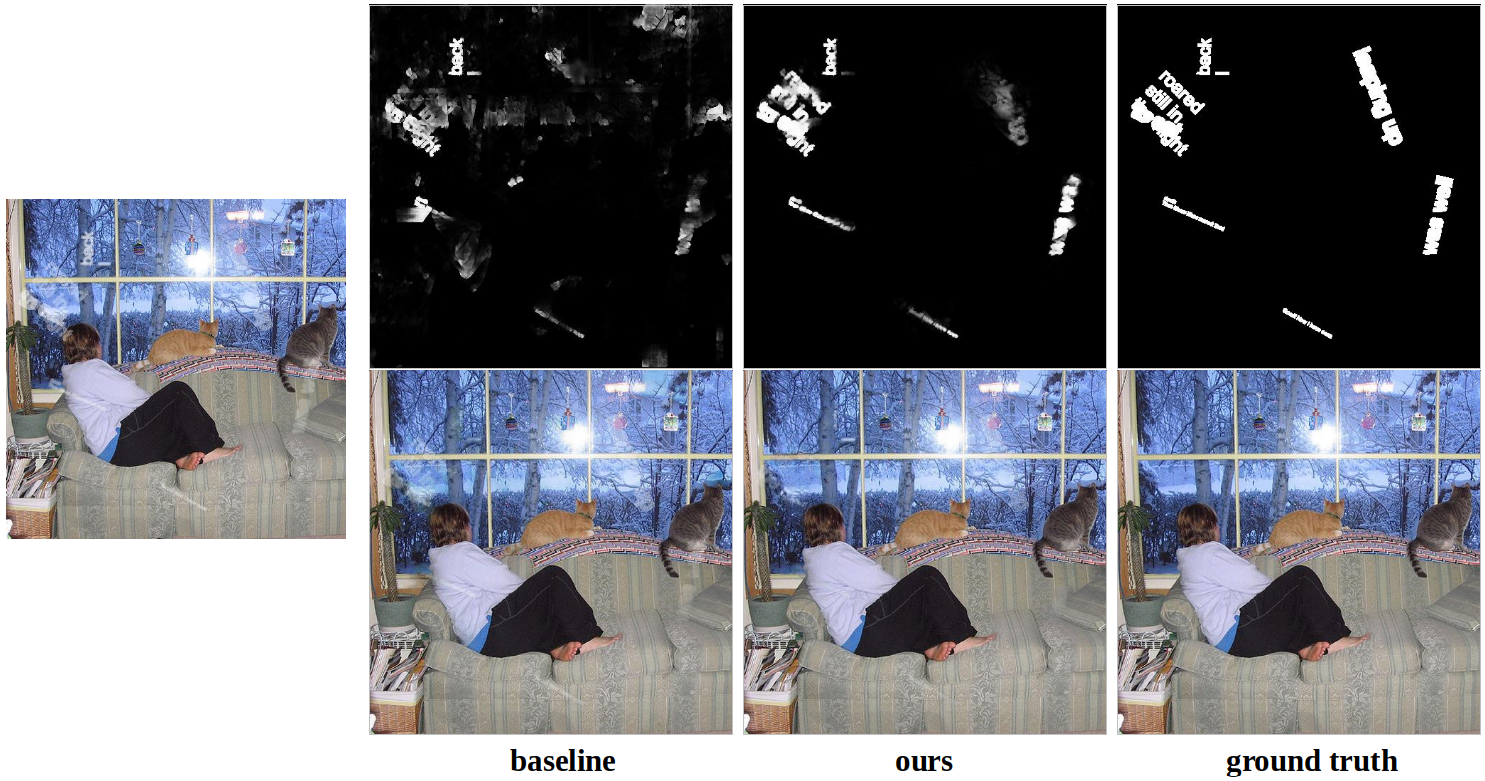}
   \caption{Predicted mask and reconstructed image comparison with baseline and ours}
   \label{fig:onecol}
\end{figure}

\begin{figure}[t]
  \centering
   \includegraphics[width=0.7\linewidth]{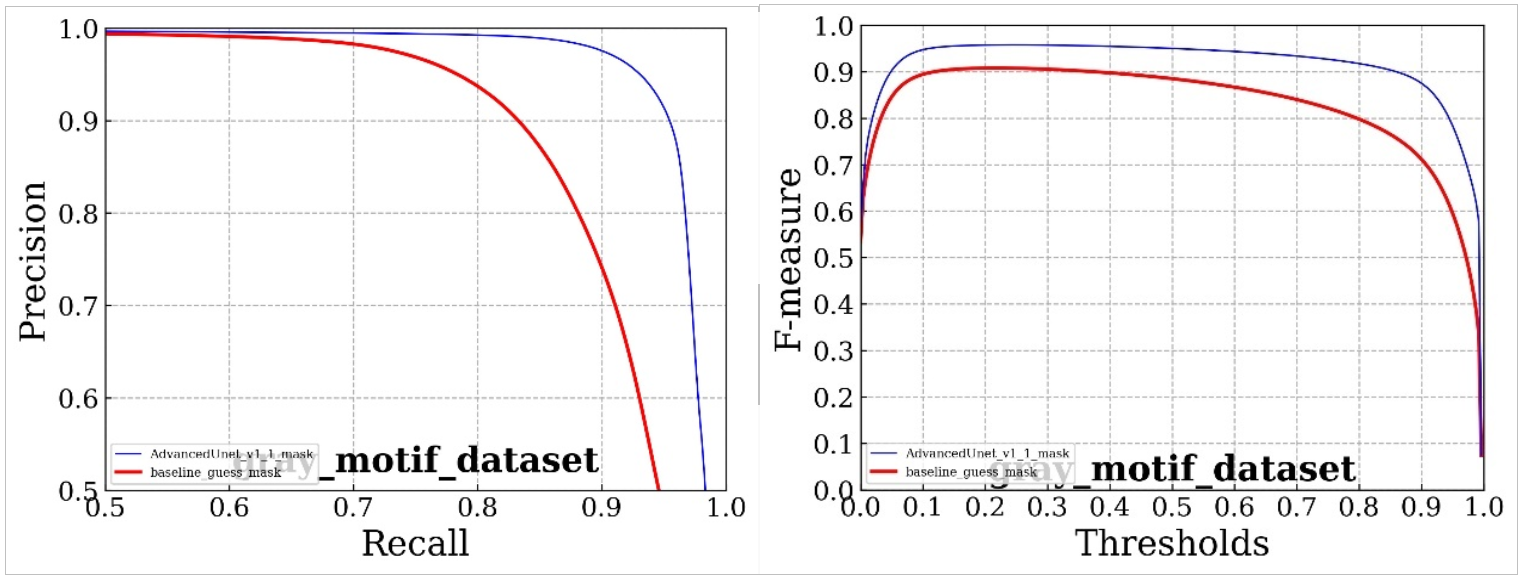}
   \caption{Quantitative analysis comparison}
   \label{fig:onecol}
\end{figure}
\subsubsection{Quantitative analysis}
In terms of specific quantitative metrics (as shown in Table 1), the average absolute error of our model in predicting the binary watermarked image relative to the real binary watermarked image is 0.010, which is 0.013 values down from 0.023 in the original model, but in percentage terms it is 56\% optimized. Similarly, in terms of watermark binary mask prediction, we can clearly see from the mIoU metric that our model achieves excellent results in watermark binary mask prediction, and we calculate the structural similarity (SSIM) of the predicted binary watermark mask with respect to the real watermark mask, and from the above data, we can see that our model also outputs a binary mask with fewer artifacts and smoother edges. We also calculated the structural similarity index and PSNR of the reconstructed image relative to the original image without watermark, and achieved a 1\% and 42\% improvement relative to the original model, respectively. 
\section{Conclusion}
We have proven the effectiveness of our method through a series of experiments, in this paper, we propose a one-stage neural network named AdvancedUnet to extract and remove the visual watermark simultaneously, and surpass the baseline by a large margin. The AdvancedUnet with RSU has a strong capability of multi-scale pixel-wise feature extraction, and the novel loss function also helps a lot with generating a smoother reconstructed image. Besides blind watermark removal, our method could also be applied to other related tasks, such as shadow removal, and reflection removal in future work.

{\small
\bibliographystyle{unsrt}
\bibliography{egbib}
}

\end{document}